\documentclass[letterpaper, 10 pt, conference]{ieeeconf}  

\IEEEoverridecommandlockouts                              

\usepackage{graphics} 
\usepackage{epsfig} 

\usepackage{balance}

\title{\LARGE \bf
Hypothesis on the Functional Advantages of the Selection-Broadcast Cycle Structure: Global Workspace Theory and Dealing with a Real-Time World
}

\author{Junya Nakanishi$^{1}$, Jun Baba$^{2}$, Yuichiro Yoshikawa$^{1}$, Hiroko Kamide$^{3}$, and Hiroshi Ishiguro$^{1}$ 
\thanks{*This work was not supported by any organization}
\thanks{$^{1}$Junya Nakanishi, Yuichiro Yoshikawa, and Hiroshi Ishiguro are with the Graduate School of Engineering Science, The University of Osaka, Osaka 560-0043, Japan
        {\tt\small nakanishi.junya@irl.sys.es.osaka-u.ac.jp}}%
\thanks{$^{2}$Jun Baba is with AI Lab, CyberAgent Inc., Tokyo 150-0042, Japan}%
\thanks{$^{3}$Hiroko Kamide is with Faculty/School of Law, Kyoto University, Kyoto 606-8501, Japan}%
}

\begin{document}

\maketitle
\thispagestyle{empty}
\pagestyle{empty}

\begin{abstract}

This paper discusses the functional advantages of the Selection-Broadcast Cycle structure proposed by Global Workspace Theory (GWT), inspired by human consciousness, particularly focusing on its applicability to artificial intelligence and robotics in dynamic, real-time scenarios. While previous studies often examined the Selection and Broadcast processes independently, this research emphasizes their combined cyclic structure and the resulting benefits for real-time cognitive systems.
Specifically, the paper identifies three primary benefits: Dynamic Thinking Adaptation, Experience-Based Adaptation, and Immediate Real-Time Adaptation. This work highlights GWT's potential as a cognitive architecture suitable for sophisticated decision-making and adaptive performance in unsupervised, dynamic environments. It suggests new directions for the development and implementation of robust, general-purpose AI and robotics systems capable of managing complex, real-world tasks.

\end{abstract}

\section{INTRODUCTION}

In recent years, a major research theme in the fields of artificial intelligence (AI), robotics, and cognitive science has been how to implement the advanced intelligence and flexible problem-solving abilities of humans and animals into systems~\cite{hassabis2017neuroscience,ho2022cognitive}. With the technical advances in machine learning (most notably deep learning) and the heightened performance of hardware in robotics, there has been growing interest in “multimodal” and “parallel” architectures that carry out tasks while simultaneously leveraging multiple cognitive functions~\cite{kotseruba202040,ajay2023compositional}. However, even if several specialized modules (e.g., vision, language, logical reasoning, motor control) each have excellent capabilities, there are still many aspects of information exchange and control methods that have not been fully organized due to the simultaneous parallel operation of multiple modules~\cite{liu2025advances}. 

Against this background, the Global Workspace Theory (GWT), which was devised by imitating human consciousness, is attracting attention. GWT positions “consciousness” from the perspective of information processing structure and proposes a framework in which information that has been competed for and integrated among numerous parallel specialized modules is temporarily brought “into consciousness” and then shared system-wide~\cite{baars2005global}. Since it was first proposed by the psychologist Bernard Baars, GWT has been linked to many empirical findings in neuroscience and cognitive science~\cite{dehaene2001towards,mashour2020conscious}. More recently, its advantages as an information processing architecture have begun to attract attention in AI research as well. Previous GWT research suggests that the “Selection” process, which integrates information among multiple parallel specialized modules, and the “Broadcast” process, which disseminates the selected information throughout the system, are expected to be effective as a wide range of functions, including creative thinking, transfer learning, top-down control, and attention allocation~\cite{mashour2020conscious,juliani2022on,vanrullen2021deep}. However, in many of these discussions, “Selection” and “Broadcast” are treated separately, and the effectiveness of these two processes being executed in parallel and intermittently are not fully addressed.

In this paper, we call the process of exchanging information through “Selection” and “Broadcast” the “Selection-Broadcast Cycle”, and focus on it. In the Selection-Broadcast Cycle, we are considering information processing that has a time dimension, that is, information processing that is not a single information processing, but rather a series of multiple information processes, such as responding to an environment that changes over time or taking time to search for an answer. These information processing methods with a time dimension are important research topics in robotics, where real-time processing is required, and in artificial intelligence systems that handle complex tasks that require long-term learning and adaptation~\cite{LESORT202052,shaheen2022continual}. For instance, for continuous tasks that span a period of time, a robot will inevitably need to change its approach during interactions with humans. Moreover, sensor data are updated moment by moment, and task goals or external conditions may change depending on the situation. Therefore, there is a need for a real-time processing framework that can dynamically decide “when and which module to call upon” in an online setting and swiftly reflect the results in the next step.

Accordingly, this paper focuses on the Selection-Broadcast Cycle structure proposed in GWT and discusses the functional advantages its dynamic, cyclic structure offers from the perspective of applying it to the design of real AI and robotic systems. Specifically, we highlight:
\begin{description}
  \item[Dynamic Thinking Adaptation:]\mbox{}\\ a capacity to dynamically rearrange module execution order, thereby enabling flexible adaptation to unexpected task changes or evolving goals
  \item[Experience-Based Adaptation:]\mbox{}\\ an acceleration of consciousness processing by exploiting past experiences stored in memory modules, facilitating faster predictions and decision-making
  \item[Immediate Real-Time Adaptation:]\mbox{}\\ a quick intervention route to consciousness processing allows for immediate response to real-time changes
\end{description} 
Our aim is to theoretically clarify “why such a structure is useful for real-time intelligent systems.” By doing so, we hope to offer fresh insights into the design philosophy and implementation guidelines of cognitive architectures based on GWT and contribute to the development of robust, general-purpose AI and robotic systems capable of adapting to complex tasks and unknown environments.

\section{LITERATURE REVIEW}
\subsection{Overview of GWT}

\begin{figure}
\centering
\includegraphics[width=0.9\columnwidth]{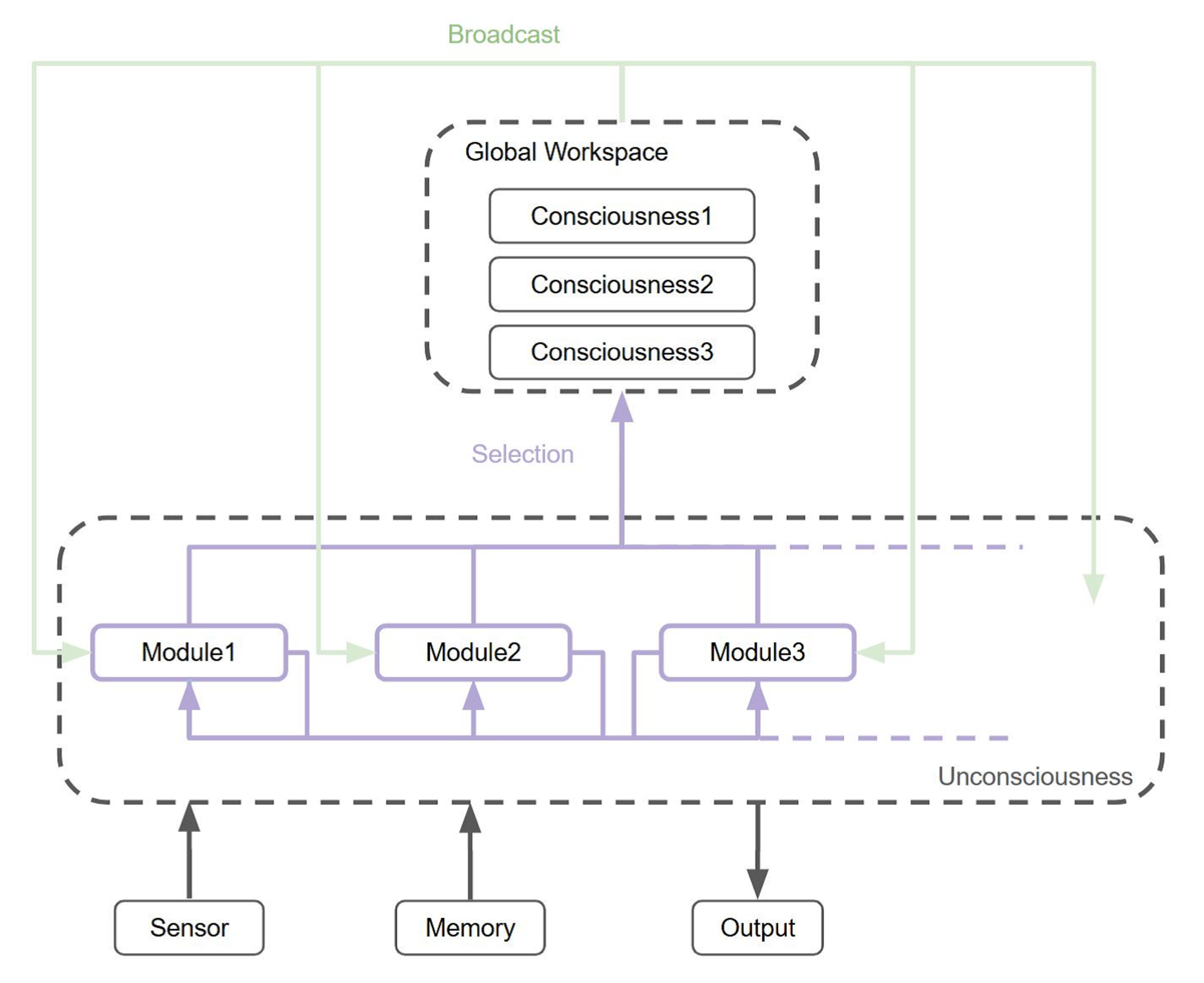}
\caption{Architecture of the Global Workspace Theory}~\label{fig:gwt}
\end{figure}

The Global Workspace Theory (GWT) is a cognitive science theory of information processing in consciousness, proposed by the psychologist Bernard Baars~\cite{baars2005global}. The essence of GWT is a framework in which information is competed and integrated among many specialized modules (e.g., vision, hearing, memory, language) that operate in parallel, and the information that eventually wins is then shared among all modules (Figure~\ref{fig:gwt}). The winning information is temporarily retained in a conscious form within a memory area called the “global workspace”. Only a limited amount of information can win at a time, and other competing information is considered to be processed unconsciously in the background. In this way, GWT is positioned as a framework to explain the interaction between a serial, limited-capacity conscious process and parallel, large-capacity unconscious processes. This model is supported by numerous experimental findings~\cite{dehaene2001towards,mashour2020conscious}. For example, in brain imaging studies (e.g., fMRI, PET, EEG), stimuli processed under consciousness involve extensive regions of the brain, including the frontal and parietal lobes, exhibiting recurrent signaling, whereas stimuli not reaching conscious awareness (i.e., unconscious processing) remain confined to local, transient activity~\cite{dehaene2001cerebral,gaillard2009converging}. This is consistent with the mechanism proposed by GWT that once some piece of information wins, it is broadcast globally to the entire system. 

On the other hand, GWT mainly deals with “What information processing structures do we use?”, so it does not provide a direct answer to the question of “Why did we arrive at this kind of information processing structure?”. From the biological and evolutionary perspective, we can address this question by considering how such a structure might have provided adaptive advantages in terms of survival and reproduction~\cite{juliani2022on}. In previous research, the focus has often been placed on the part of GWT’s information processing structure related to competing and integrating information among multiple specialized modules operating in parallel (Selection process) and on the part that shares the selected information with the entire system (Broadcast process), and the advantages and benefits of these have been discussed.

\subsection{Functional Advantages of Selection}
In this paper, the process of selecting information from among the information processed in parallel by multiple specialized modules and then integrating them in a global workspace is called the “Selection” process.

\subsubsection{Diverse Perspectives}
By comparing and examining the outputs of multiple specialized modules, it is thought that it will be possible to generate a wider variety of solutions and ideas for a given task~\cite{ito2023investigating,wiggins2012crossing}. For instance, if both a visual module and a language module are operating simultaneously, approaches that capture a problem from a pictorial/imaginative viewpoint can be compared with those that capture it from a linguistic/logical viewpoint. This concept is akin to the notion of “ensemble learning”~\cite{polikar2012ensemble} : by combining multiple models or modules with different specializations, they can complement the diverse aspects that a single model alone would not capture, thereby producing higher predictive accuracy and robustness overall. 

Furthermore, the mechanism that integrates multiple parallel modules enables unexpected combinations of knowledge and skills from each module, which is thought to lead to creative thinking~\cite{vanrullen2021deep,wiggins2012crossing}. For example, imagine a module responsible for visual thinking, inspired by metaphorical expressions provided by a language processing module, giving rise to a new diagram or prototype, which is then validated by a logical reasoning module. Alternatively, a module specializing in reinforcement learning might combine with a sensorimotor module’s proposed action strategy, leading to previously unanticipated solutions or task-execution procedures. The process of generating these incidental or divergent ideas and then evaluating, narrowing down, and integrating them is considered by many to be at the core of creative thinking~\cite{stojanov2013creativity}.

\subsubsection{Transfer Learning}
When faced with a new task, utilizing the skills already acquired in the specialized modules reduces the need to learn from scratch, and as a result, it is thought that the efficiency and speed of learning will improve~\cite{vanrullen2021deep,wiggins2012crossing}. For instance, if there are modules that excel in visual recognition, language processing, or logical reasoning and each is independently trained, then when facing a new domain or a different task, it becomes possible to adapt quickly by making use of the knowledge and representations already accumulated in these modules. This is analogous to “transfer learning”~\cite{tan2018survey} in machine learning. In fact, when adapting a deep neural network learned in one domain (source domain) to another domain (target domain), reusing the lower-level feature extraction parts shortens the early training phase while still delivering high performance.

\subsection{Functional Advantages of Broadcast}
In this paper, the process of sharing selected information with all specialized modules is called the “Broadcast” process.

\subsubsection{Shared Attention}
It is thought that broadcasting allows each specialized module to concentrate its resources on information that is deemed to be extremely important according to the current goals and environmental conditions, thereby improving the efficiency and accuracy of task execution~\cite{wiggins2012crossing,dossa2024design}. For example, consider a robot endowed with multiple sensory modules for vision, hearing, and touch, which is tasked with detecting, identifying, and accurately grasping an object. First, the visual module, operating unconsciously, generates multiple candidates, performing tasks such as location estimation and object classification in parallel. Meanwhile, the hearing module tries to gather hints from environmental sounds or voice commands that could modify actions. The tactile module prepares feedback control for the stage at which the robot actually grasps the object. After the information generated by each module is integrated by the Selection process, if the decision “to combine accurate location estimation from the visual module with minor corrective commands from auditory instructions” wins, that information is shared with all modules via the Broadcast function. As a result, the robot can carry out the plan “move the arm toward the coordinates estimated visually, corrected by auditory information” in coordination across all modules. 
 
This mechanism seems to be highly relevant to the “Transformer architecture”~\cite{vaswani2017attention}. Transformers, which demonstrate extremely high performance in various tasks such as natural language processing and image recognition, have a core mechanism known as “self-attention”. In self-attention, the inputs (or feature vectors) compute their mutual relevance, enabling the network as a whole to incorporate necessary contextual information. This mechanism is akin to GWT’s claim of handling diverse information while spotlighting important items and sharing them throughout the system. Though the transformer was not initially designed with the goal of mimicking consciousness, the fact that it achieves such high performance in language processing, image recognition, and more by way of sharing of important information hints at the fundamental usefulness of a strategy that shares the most crucial elements globally in an intelligent system.

\subsubsection{Predictive Coding}
Among the specialized modules, there are those that receive data from sensors (e.g., visual, auditory, tactile). If they receive predictions or metacognition as broadcast information, it may enhance the performance of the module’s output~\cite{vanrullen2021deep,wiggins2012crossing}. For example, when the visual module is only processing lower-level features such as raw pixel data and edge information, it will only output tentative recognition results based on local statistics and pattern recognition. However, when higher-level context and objectives such as “this scene is outdoors and there is a high possibility that there are multiple people in the picture” and “the task is to judge the facial expressions of specific people” are broadcast from the global workspace, the visual module will re-evaluate its output while referring to these predictions and hypotheses. As a result, corrections such as prioritizing the extraction of resolution and regions of interest that are appropriate for the task, or more carefully searching for clues to separate people and backgrounds, can be expected to improve recognition performance and reduce false positives. This aligns closely with the concept of “predictive coding”~\cite{friston2009predictive} often discussed in neuroscience and cognitive science. Predictive coding posits that the brain or cognitive system is constantly sending top-down predictions from higher (i.e., more advanced) modules to lower (i.e., more basic) modules, while the lower-level modules calculate and return the discrepancy (prediction error) between the actual sensory input and the prediction. If the discrepancy is large, it implies that something different from the predictions is likely in the scene, and this error is returned upstream so that the higher-level module can update or generate new predictions. If the discrepancy is small, it implies that the prediction and actual data largely match, thus increasing the likelihood that it is really as observed. Through repeated mutual interplay between top-down predictions and bottom-up prediction errors, the entire perception and cognition system dynamically adapts to the environment.

\section{HYPOTHESIS}

In this paper, in addition to the structural advantages from each of the traditional GWT perspectives (Selection and Broadcast), we newly focus on the advantage of a cycle structure in which information processing occurs through Selection and Broadcast (Selection-Broadcast Cycle). Within this cycle structure, we discuss the dynamic, stepwise information processing in which Selection and Broadcast intertwine in parallel and intermittently.

\subsection{Dynamic Thinking Adaptation}
The Selection-Broadcast Cycle possesses a structure that can realize any order of serial processing steps of specialized modules. The serial processing referred to here means processing that is carried out step by step (e.g., a chain of thought~\cite{wei2022chain}, inductive and deductive reasoning~\cite{shanahan2022deductive}). In contrast to parallel processing, in which multiple modules operate simultaneously, serial processing involves processing being carried out in order, with the information generated or selected by one module being passed on as input to the next module. In serial processing, the final answer is derived from the inferences and logical development that take place in the intermediate processing. This process of deriving conclusions in steps allows reliable problem solving and decision making with a small number of inferences and logical knowledge for various complex tasks. For example, by simply memorizing the results of addition and multiplication of 0 to 9 and the methodology of longhand arithmetic, you can calculate any addition or multiplication of integers (e.g., 11×2=10×2+1×2). In this way, by breaking down complex tasks into simpler sub-tasks (i.e., tasks that can be processed using limited memory or simple rules) and dealing with them in stages, it is possible to deal with a wide range of different tasks using relatively little memory capacity.

\begin{figure}
\centering
\includegraphics[width=0.5\columnwidth]{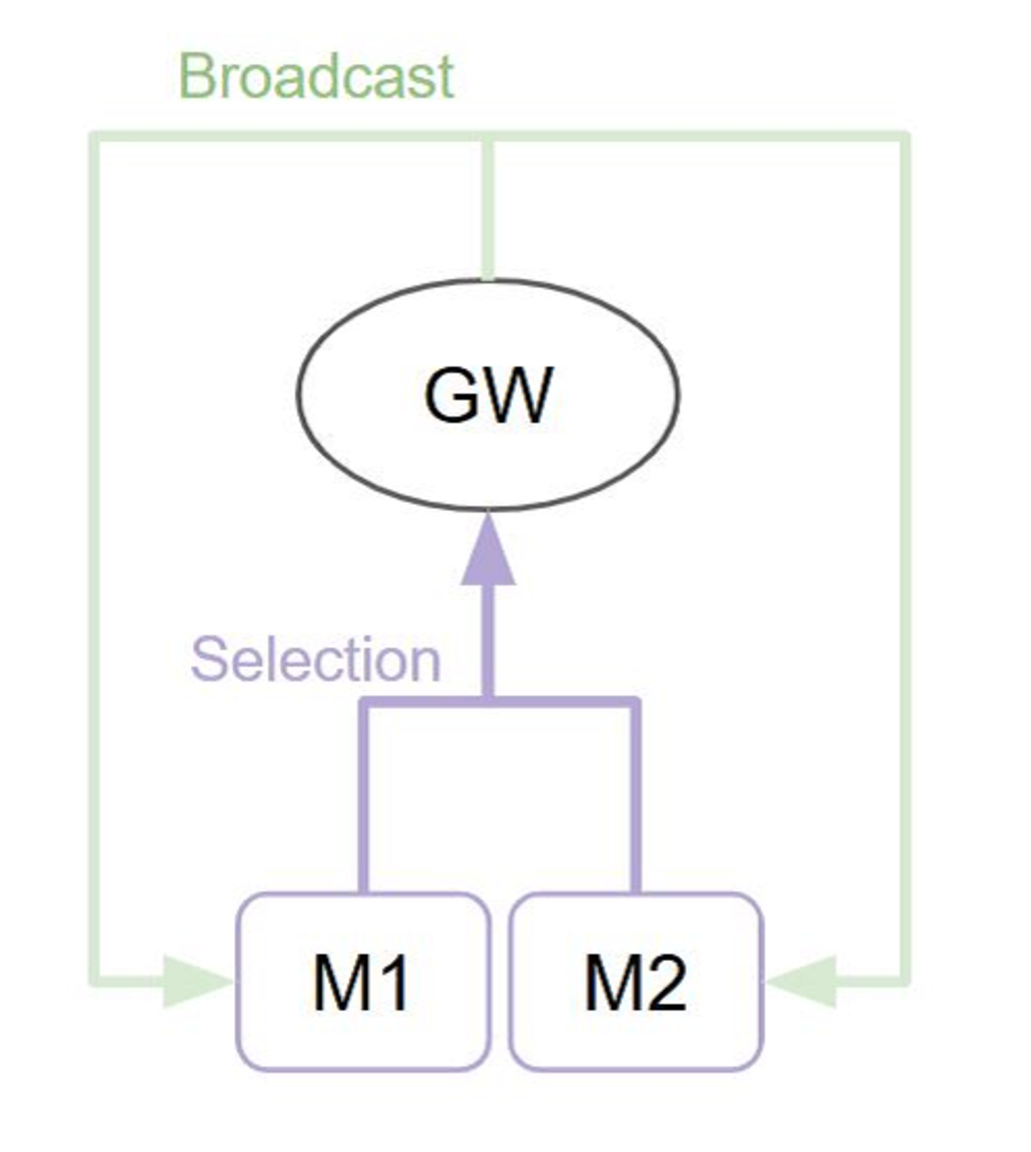}
\caption{Example of GWT-based structure with two modules}~\label{fig:example}
\end{figure}

\begin{figure}
\centering
\includegraphics[width=0.99\columnwidth]{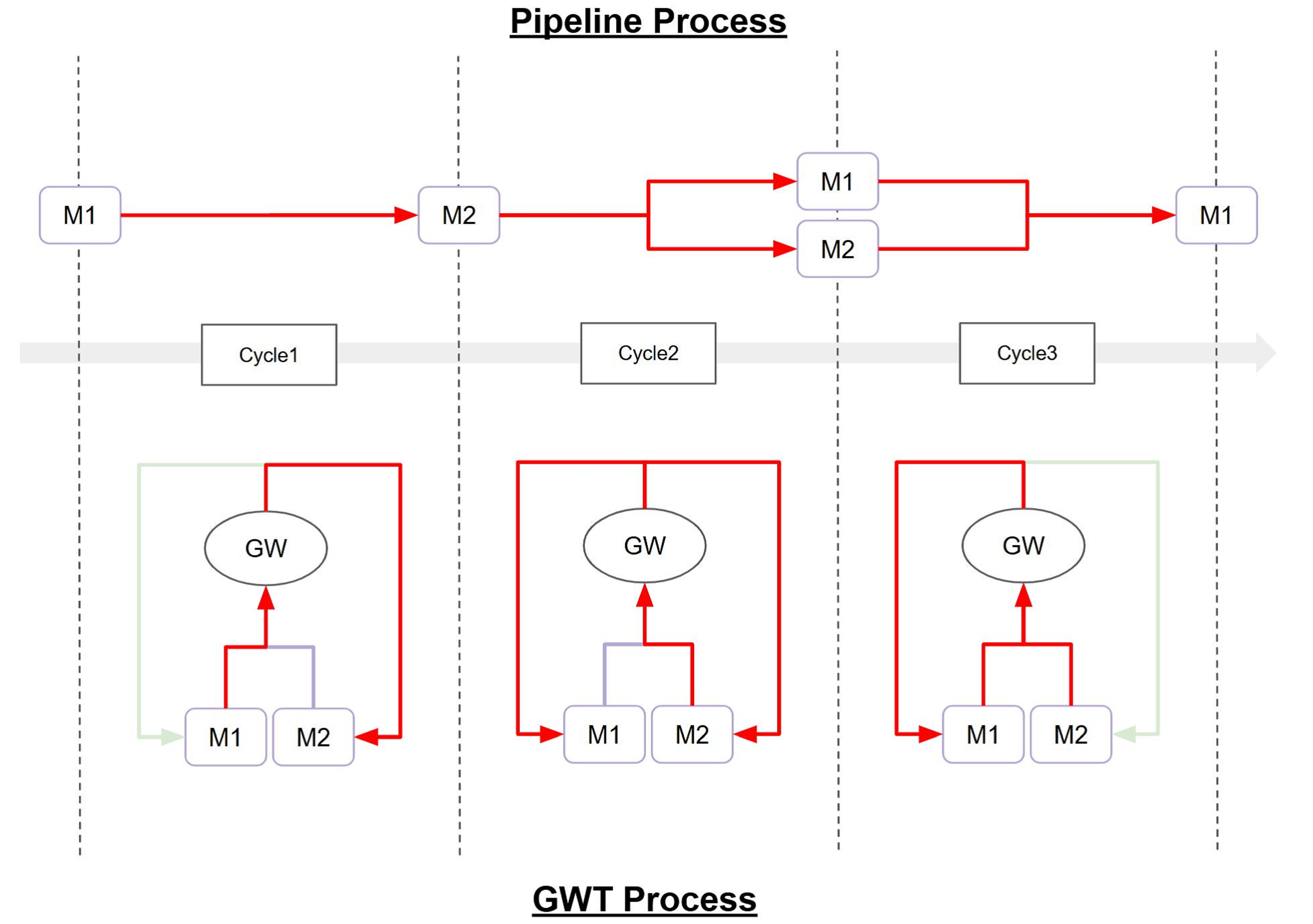}
\caption{Flow of pipeline and GWT process in the GWT-based example}~\label{fig:changarable}
\end{figure}

The Selection-Broadcast Cycle process has a space where such intermediate inferences and logical developments can be freely performed. Figure~\ref{fig:example} shows an example of a simple Selection-Broadcast Cycle structure with two modules (M1, M2). The upper part of Figure~\ref{fig:changarable} shows an example of the execution procedure of modules, and the lower part shows the processing flow that executes that execution procedure in the Selection-Broadcast Cycle. As you can see, the Selection-Broadcast Cycle process can execute any execution procedure using the modules by switching the selection well. In order to implement such a vast serial processing space for intermediate inferences and logical development as a pipeline, it is thought that a large tree structure made up of a large number of modules is necessary. The Selection-Broadcast Cycle process is thought to be a structure made up of a minimum number of modules using looped information processing.

Furthermore, this function enables flexible and dynamic processing, allowing you to try out all kinds of thought processes and change your thought processes in response to changes in the situation. This is a great advantage when dealing with situations that are difficult to handle with a fixed pipeline process, such as when the processing procedure is unclear or the goal is changed partway through. For example, consider the case where a robot explores a room based on information from multiple sensors (vision, touch, audio input, etc.). At the start of the search, the main objective was to search for and move along the shortest route, and the processing was set up to call the object detection module and the route planning module in order. However, during the search, there were many collisions with people in the room along the route. In this case, the Selection-Broadcast Cycle makes it possible to share the problem with the whole system, devise a solution, and make changes to the processing, for example, by calling a human detection module while planning a route. Also, if a voice instruction is received and the content of the instruction changes, it is possible to call a voice recognition module to share the analysis results with the whole system, and then reconfigure the execution order of the visual module and route planning module in response to the results. Thanks to this variable serial processing, the order in which the necessary specialized modules are called can be flexibly rearranged in response to changes in the situation or new goals, making it possible to accomplish tasks that would be difficult with fixed pipeline processing.

\begin{figure*}
\centering
\includegraphics[width=1.7\columnwidth]{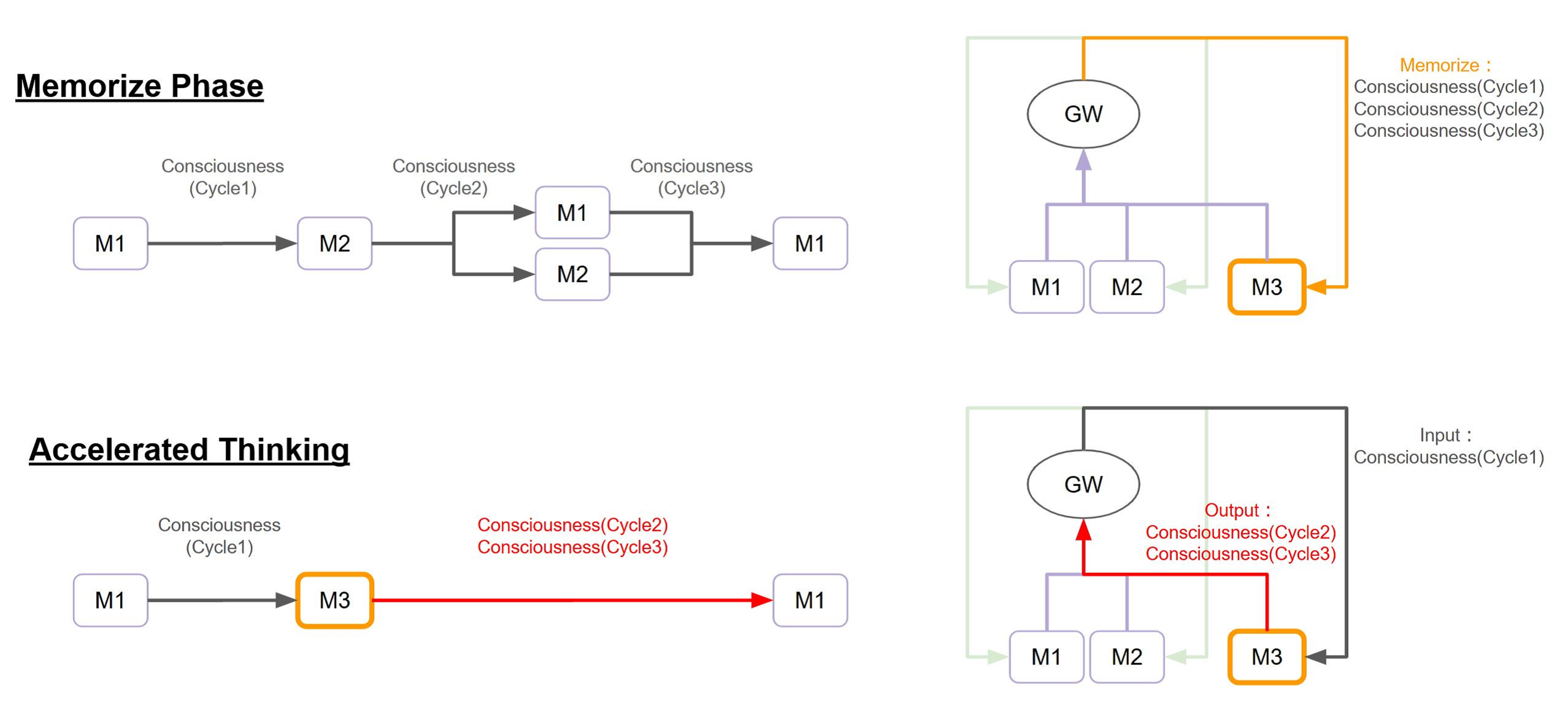}
\caption{Flow of accelerated thinking in the GWT-based example}~\label{fig:experience}
\end{figure*}

This function can also be said to mean that it is possible to exchange information between any of the modules. VanRullen and Kanai~\cite{vanrullen2021deep} point out that the global workspace functions as a “hub” between specialized modules, and that cycle-consistency learning~\cite{zhu2017unpaired} can be carried out by exchanging information between the same specialized modules. Cycle-consistency learning is a learning method that imposes constraints on the model to maintain consistency when converting data back and forth. These constraints ensure that once converted data can be restored to its original state by reversing the conversion, and prevent the loss of content or meaning during the conversion process. A major advantage is that it can learn domain mapping even without training data. In this way, the outputs of each specialized module are continuously cross-checked by repeating the Selection-Broadcast Cycle, and the entire system has the potential to detect potential inconsistencies, correct errors, and gradually build more reliable processing results.

\subsection{Experience-Based Adaptation}
As noted, in GWT, the information that is sequentially raised in the Global Workspace (Consciousness) through the Selection-Broadcast Cycle is shared with all specialized modules in a stepwise manner. Here, we focus on the point that the serial processing carried out in consciousness enters each specialized module in chronological order. It is thought that there are specialized modules that record such chronological consciousness and store it as experience memory~\cite{franklin2005role}. We can further suppose that such experience memory can be recalled if a similar situation arises. If so, it would become possible to speed up or predict the course of serial processing.

Figure~\ref{fig:experience} shows an example of a simple Selection-Broadcast Cycle structure with two modules (M1, M2) and one experience memory module (M3). Consciousness(Cycle1), Consciousness(Cycle2), and Consciousness(Cycle3) are each listed in the global workspace in chronological order. Since these consciousnesses are broadcast in chronological order, they of course flow into the experience memory module as well. The experience memory module retains them as experiences. Then, when Consciousness(Cycle1) is broadcast again, the experience memory module can output Consciousness(Cycle2) and Consciousness(Cycle3) as recalled memories. This means that it is possible to reach the output of Consciousness(Cycle3) in two cycles, whereas it would have taken three cycles to reach it in the past. As described above, it is thought that the Selection-Broadcast Cycle will enable faster serial processing and prediction.
This is similar to the concept of “chunking”~\cite{gobet2001chunking} in cognitive science, and if learned schemas and procedures are stored as a kind of “chunk”, then when faced with a similar task next time, that chunk can be called up all at once to quickly progress with the processing. 

This mechanism not only increases processing speed, but also promotes inference and anticipation of actions. In other words, while referring to past thought processes, it is possible to make predictions such as “there is a possibility that new information will be lacking at this stage” or “it would be better to activate the sensorimotor module before the logical inference module in the next step”, and it is possible to adjust the order of module calls and resource allocation in advance based on these predictions.
As a result, each step in the variable serial processing is no longer a simple “trial and error” process, but rather a planned and efficient process that makes full use of past accumulated knowledge. The meta-cognitive decisions made during this process, such as “which module should be activated at what time” and “when should top-down information be updated”, are also optimized through the use of overall information sharing and memory via the Selection-Broadcast Cycle. In this way, by having a system in place that can record and utilize a record of serial processing, it is hoped that the cognitive architecture based on GWT will not only speed up, but also acquire advanced problem-solving capabilities that incorporate reasoning and prediction with an eye on the next move.

There have been several implementations of agent systems that apply experience memory as knowledge (e.g., reasoning and prediction)~\cite{Laird2012Cognitive,martin2021affective}. For instance, Franklin and colleagues~\cite{franklin2013lida} have demonstrated a framework called LIDA (Learning Intelligent Distribution Agent), which builds on GWT to incorporate conscious content into various cognitive modules, including an episodic memory module. In LIDA-based implementations, information that reaches consciousness is not only broadcast to specialized modules but is also chronologically recorded in an episodic (or experience) memory. When a similar situation occurs, the system recalls the sequence of recorded conscious events and applies them as learned knowledge.

\begin{figure}
\centering
\includegraphics[width=0.99\columnwidth]{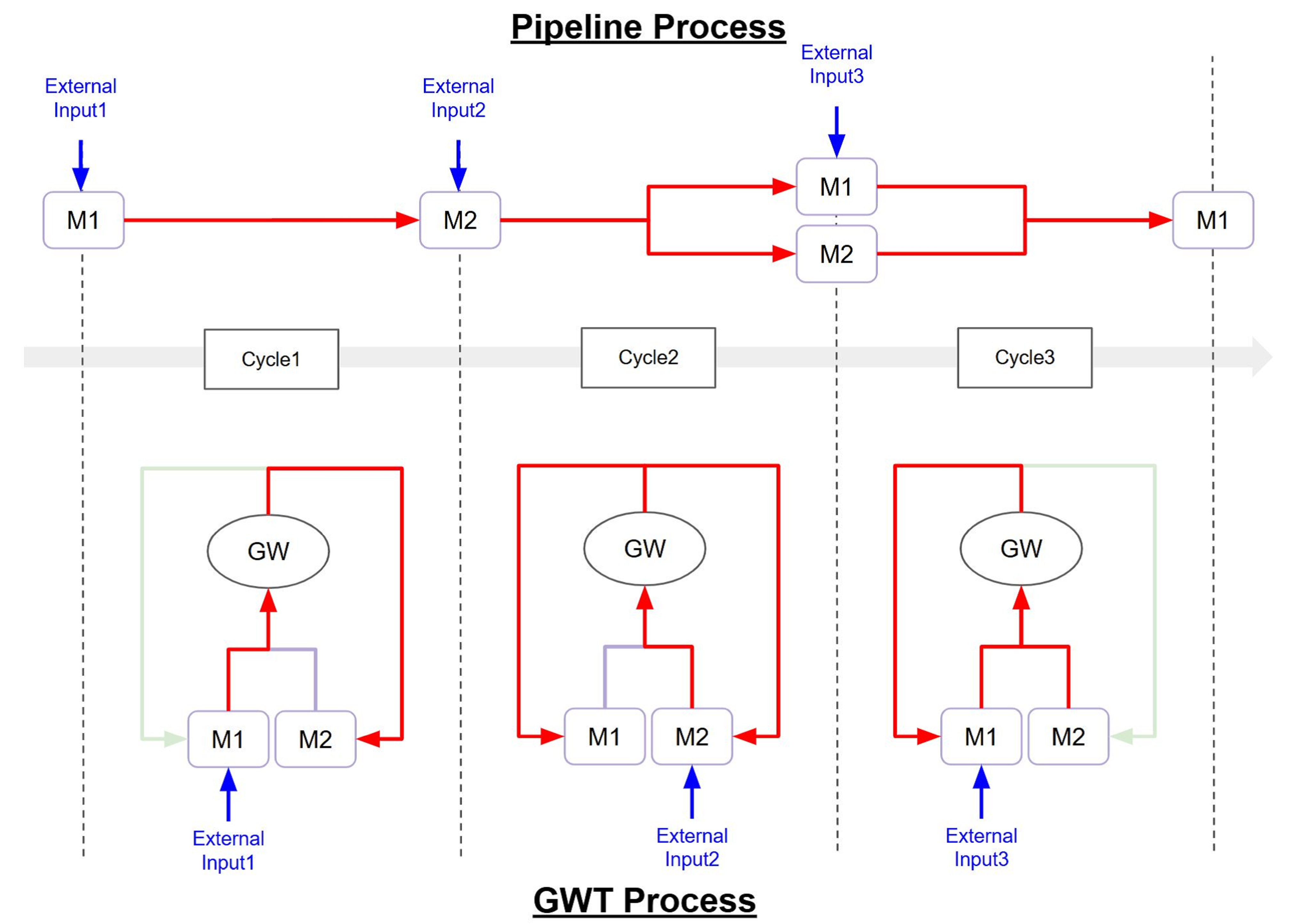}
\caption{Flow of real-time intervention in the GWT-based example}~\label{fig:reactive}
\end{figure}

\subsection{Immediate Real-Time Adaptation} 
The selective broadcast cycle allows real-time intervention in the results of intermediate processing by external input. Figure~\ref{fig:reactive} shows a simple scenario in which external intervention occurs within a Selection-Broadcast Cycle process with two modules (M1, M2). As shown, external inputs can influence the global workspace's serial processing at any point. For example, if a specialized module detects new, highly significant information, this information can immediately enter the global workspace through the Selection process, which then disseminates it to all other modules via Broadcast. This quick route ensures that unnecessary waiting times and message passing are greatly reduced, and real-time system responsiveness is greatly improved.

In practical robotics scenarios, such flexible intervention mechanisms have notable advantages. For instance, imagine a robot performing an assembly task using multiple sensory modules (visual, tactile, auditory). Suppose the robot's tactile sensor suddenly detects an unexpected slip or instability in its grip. With the Selection-Broadcast Cycle, this critical information is rapidly promoted into the global workspace, interrupting the ongoing processing sequence. Consequently, other modules (e.g., motor control, vision processing, or reinforcement learning) immediately receive this alert and can swiftly initiate corrective actions. This immediate broadcast enables the system to promptly reconsider and revise its gripping strategy from both top-down (strategic re-planning) and bottom-up (sensor-driven adjustments) perspectives, substantially improving safety, precision, and robustness in real-time.

\section{DISCUSSION}
Traditional discussions of GWT’s intelligence have predominantly emphasized the process on static, supervised settings, which rely heavily on pre-labeled data sets, explicit instructions, and predefined tasks (e.g., ensemble learning, transfer learning, self-attention, predictive coding). In such scenarios, intelligence manifests primarily as a system’s ability to accurately replicate patterns and knowledge derived from historical, structured data. However, the real-world application of artificial intelligence increasingly demands a shift toward dynamic, unsupervised settings, where tasks, environments, and goals continuously evolve, often without explicit guidance or labeled examples.

In dynamic, unsupervised scenarios, intelligent systems face fundamentally different challenges. Rather than relying on historical labels or fixed benchmarks, these systems must autonomously discover meaningful patterns, adapt swiftly to changing contexts, and continuously learn from ongoing experiences. In this paper, we discussed the strengths of GWT in such real-time processing by focusing on Selection-Broadcast Cycle. We explained that this Selection-Broadcast Cycle realizes flexible processing, is capable of being accelerated, and is a mechanism that can respond immediately to real-time changes. Thus, by highlighting the advantages of the Selection-Broadcast Cycle, this paper extends traditional conceptions of GWT intelligence into the realm of dynamic, unsupervised learning, opening new pathways toward the development of more robust, adaptive, and autonomous artificial intelligence systems capable of thriving in complex real-time environments. Future research could further explore practical implementations and empirical evaluations to validate these theoretical insights and expand the applicability of GWT-based architectures in diverse, real-world scenarios.

Furthermore, although GWT seems well-suited for thriving in the real-time world, one potential way to enhance its adaptability further could involve multiple consciousness (GWT) processes operating in parallel. This parallelization could facilitate the simultaneous exploration of diverse solutions, enhance adaptability by rapidly responding to varied and unpredictable changes, and effectively distribute cognitive load, thereby potentially surpassing the limitations inherent to a single, centralized consciousness structure. Such a mechanism might represent the collective intelligence observed in groups of humans, suggesting that human societies themselves could represent natural exemplars of parallel consciousness networks capable of robust, adaptive decision-making in complex and dynamic environments. For example, Taniguchi~\cite{taniguchi2024collective} is researching the dynamics of such group intelligence and language development.


\section{LIMITATIONS AND FUTURE WORK}

While the proposed Selection-Broadcast Cycle structure inspired by the Global Workspace Theory (GWT) provides a compelling theoretical framework for adaptive, real-time cognitive architectures, several critical limitations need to be acknowledged and addressed in future work.

One significant limitation of this study is the absence of empirical validation. The advantages of the Selection-Broadcast Cycle, such as dynamic thinking, experience-based acceleration, and immediate real-time responsiveness, remain largely theoretical. Currently, the paper does not present experimental results, simulations, or quantitative analyses to substantiate these claims. Therefore, readers must accept the described benefits without direct evidence of improved adaptability or efficiency compared to other existing methods. To strengthen future iterations of this research, practical implementations such as comparative simulations or robot-based experiments demonstrating fewer task failures or quicker adaptation would be essential. 

In particular, the overall effectiveness of the system depends greatly on the quality of the Selection process. There remain important open questions regarding whether such effective and sophisticated selection mechanisms can actually be put to practical use. In static environments, there is an example of Selection process that improves the overall performance of the system by weighting based mainly on internal indicators such as happiness and past experience~\cite{garrido2022global}. In order to put GWT-based architectures to practical use in real-world applications, it is extremely important to achieve robust and adaptive Selection processes, so future research will need to address this implementation issue in dynamic environments as well.

\section{CONCLUSIONS}

In this paper, we explored the potential of the Global Workspace Theory (GWT) and, in particular, the Selection-Broadcast Cycle, as an information processing architecture suitable for dynamic, unsupervised real-time environments. Traditional approaches to artificial intelligence often rely heavily on structured, labeled data, where intelligence primarily involves replicating known patterns. However, real-world applications require systems that can continuously adapt and respond to evolving tasks, environments, and goals. In this context, we highlighted the Selection-Broadcast Cycle’s strengths: its flexibility to rearrange module execution order dynamically, its capability for acceleration through experience-driven predictions, and its responsiveness to immediate real-time inputs.

Our hypothesis suggests that a cognitive architecture based on GWT and, specifically, the Selection-Broadcast Cycle, provides a robust framework for dynamic decision-making and rapid adaptation in complex environments. The ability to rearrange processing sequences dynamically, accelerate learning through experience-based memory, and intervene swiftly in response to changing conditions positions GWT-based architectures to effectively handle the challenges posed by real-time intelligence.


A critical unresolved question remains the practical feasibility of implementing robust and adaptive Selection mechanisms in real-world systems. Future research must address this challenge, potentially through integrating machine learning techniques and advanced evaluative frameworks, to further validate and extend the applicability of GWT-based architectures. By tackling these challenges, we can move closer to developing truly autonomous, flexible artificial intelligence systems capable of thriving in the complexities and uncertainties of the real-time world.

\addtolength{\textheight}{-12cm}   







\balance


\bibliographystyle{IEEEtran}

\bibliography{root}


\end{document}